\documentclass[10pt]{article}

\usepackage[preprint]{neurips_2025}
\PassOptionsToPackage{numbers,square,comma}{natbib}
\usepackage{times}
\usepackage{epsfig}
\usepackage{graphicx}
\graphicspath{{Figures/}}
\usepackage{amsmath}
\usepackage{amssymb}
\usepackage{caption}
\usepackage{subcaption}
\usepackage{multirow}
\usepackage[colorlinks=true, urlcolor=blue, linkcolor=red]{hyperref}
\usepackage[ruled,linesnumbered]{algorithm2e}
\usepackage[accsupp]{axessibility}  
\usepackage{microtype}      
\usepackage{booktabs}       
\usepackage{nicefrac}       
\usepackage{color}
\usepackage{tikz}
\usepackage{xcolor}
\usepackage{textcomp}

\title{LLM‑Guided Taxonomy and Hierarchical Uncertainty for 3D Point Cloud Active Learning}

\author{Chenxi Li$^{\dag}$, Nuo Chen$^{\dag}$, Fengyun Tan, Yantong Chen, Bochun Yuan
\\ 
\textbf{Tianrui Li}, \textbf{Chongshou Li}$^{\ddag}$ \\ 
Southwest Jiaotong University, Chengdu, China \\
\{lcx041511, chennuo2022, tfy, 2022116015, ybc23793\}@my.swjtu.edu.cn \\
\{trli, lics\}@swjtu.edu.cn \\
$^\dag$Equal Contribution, $^\ddag$Corresponding Author
}

\begin{document}

\maketitle

\begin{abstract}

We present a novel active learning framework for 3D point cloud semantic segmentation that, for the first time, integrates large language models (LLMs) to construct hierarchical label structures and guide uncertainty-based sample selection. Unlike prior methods that treat labels as flat and independent, our approach leverages LLM prompting to automatically generate multi-level semantic taxonomies and introduces a recursive uncertainty projection mechanism that propagates uncertainty across hierarchy levels. This enables spatially diverse, label-aware point selection that respects the inherent semantic structure of 3D scenes. Experiments on S3DIS and ScanNet v2 show that our method achieves up to 4\% mIoU improvement under extremely low annotation budgets (e.g., 0.02\%), substantially outperforming existing baselines. Our results highlight the untapped potential of LLMs as knowledge priors in 3D vision and establish hierarchical uncertainty modeling as a powerful paradigm for efficient point cloud annotation.  
\end{abstract}

\begin{figure}[hbtp]
  \centering
  \includegraphics[width=\textwidth]{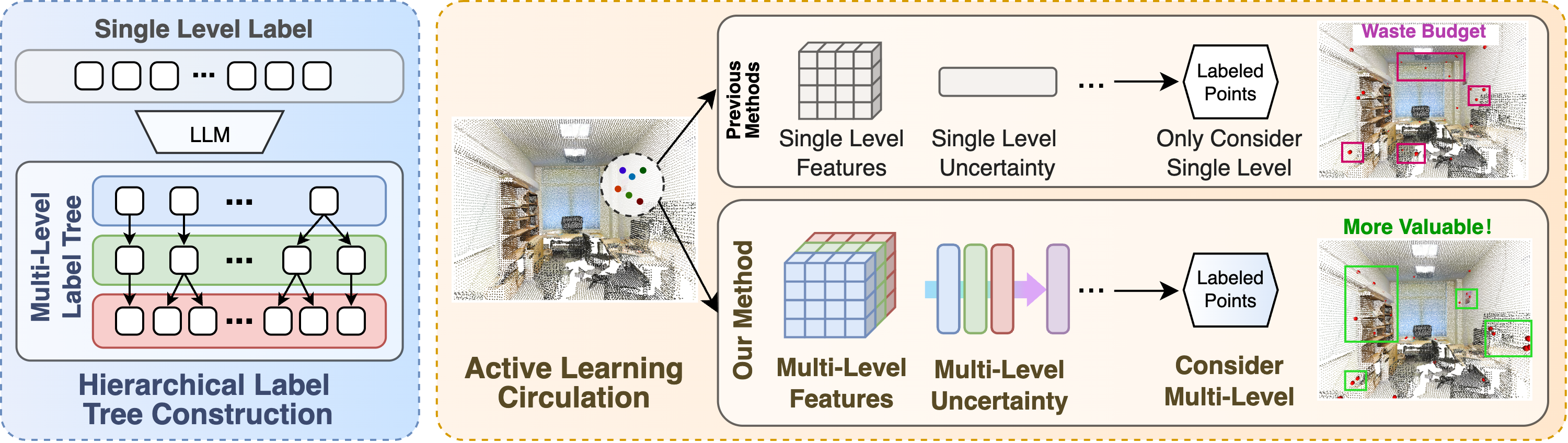}
  \caption{ 
  Overview comparing the proposed LLM-Guided hierarchical point cloud active learning and the existing single-level point cloud active learning. We propose a new recursive uncertainty propagation method.   
  }
  \label{fig:teaser}
\end{figure}

\section{Introduction}
Deep neural networks have achieved remarkable success in 3D point cloud semantic segmentation for applications like autonomous driving and robotics. However, obtaining dense point-level annotations for large 3D scans is extremely labor intensive and costly in practice \cite{choy20194d, hu2020randla}. A single indoor scene can contain hundreds of thousands of points, making full annotation prohibitively time-consuming. To alleviate this burden, researchers have explored training with fewer labels via weak supervision \cite{Liu_2021_CVPR, su2023weakly} (e.g., labeling only a small fraction of points or using coarse labels) and semi-supervision \cite{jiang2021guided, cheng2021sspc} (using unlabeled point clouds to learn from pseudolabels). Another prominent direction is active learning \cite{lin2020active, shao2022active}, which iteratively selects the most informative samples or regions to label, thus maximizing the utility of each annotation. These strategies substantially reduce the annotation requirement while maintaining competitive performance.
 
Despite these advances, most existing semi-supervised methods assume a flat set of semantic classes and ignore the inherent hierarchical structure of labels in 3D environments. In real-world scenes, object categories naturally form a semantic hierarchy (for example, indoor - furniture - chair), which aligns with human cognitive understanding of scenes. This lack of hierarchical reasoning constitutes a critical limitation: the inability to explicitly incorporate coarse-grained semantic context (e.g., object category) into fine-grained predictions can exacerbate label ambiguity and increase vulnerability to anomalous inputs. A study highlights that explicitly modeling class relationships can improve segmentation reliability \cite{liang2018dynamic}, yet no prior work integrates such semantic hierarchy information into an active or semi-supervised learning framework for point cloud segmentation.

To bridge this gap, we propose a hierarchical label-aware active learning approach for efficient 3D point cloud segmentation as shown by Figure \ref{fig:teaser}. In our framework, we leverage a predefined semantic label tree generated by the Large Language Model and introduce a cross-level uncertainty propagation mechanism, integrating it seamlessly with spatially-aware active learning to reduce redundant annotations. Uniquely, our method combines semi-supervised learning, active learning, and hierarchical semantic reasoning in a single unified model. The key contributions of this work are summarized as follows:

\begin{itemize}
    \item \textbf{Hierarchical Label Tree Construction:} We construct a multi-level semantic label hierarchy  with LLM-guided refinement, ensuring consistent and balanced clustering. To reflect semantic structures suited for point cloud segmentation, task-driven constraints — such as hierarchical balance, inter-layer differentiation, and flexible handling of ambiguous labels — are incorporated into the prompt to guide the construction of a semantically coherent label hierarchy.

\item \textbf{Hierarchical Semantic Uncertainty Modeling:} We propose a cross-level uncertainty propagation strategy that recursively aggregates uncertainty from coarse to fine levels.   This allows the model to use higher-level semantic context to resolve confusion among similar fine-grained categories, improving uncertainty estimation. Additionally, we integrate global-aware uncertainty alignment to refine point selection based on the overall uncertainty distribution.

\item \textbf{Unified Semi-Supervised and Active Learning Framework with Hierarchical Awareness:} We present the first framework combining hierarchical labels, semi-supervised learning, and active learning for 3D point cloud segmentation. Hierarchy-aware uncertainty guides active point selection toward diverse, informative regions by prioritizing points that align with both local uncertainty and global semantic distribution, thus maximizing the effectiveness of each annotation. Experiments demonstrate that the proposed framework outperforms existing semi-supervised or active learning methods.\end{itemize}

\section{Related Work}
Efforts to reduce the annotation burden in 3D semantic segmentation have led to semi-supervised and weakly supervised methods~\cite{zhang2023deep}. These approaches either sparsely label points~\cite{Liu_2021_CVPR} or propagate label information using self-training and teacher-student consistency~\cite{cheng2021sspc}, yet they still lag behind fully supervised models. Active learning offers a promising alternative by selecting informative samples under limited budgets. Early work by Lin et al.~\cite{lin2020active} partitioned scenes into blocks for sampling, while Wu et al.~\cite{wu2021redal} proposed selecting sub-regions using entropy and visual cues. More recently, Xu et al.~\cite{xu2023hierarchical} introduced a per-point selection method using hierarchical uncertainty and feature suppression, achieving near-supervised performance with just 0.1\% labeled points. However, these approaches treat class labels as flat and independent, ignoring semantic hierarchies—resulting in less reliable point selection in structurally complex scenes. Our work addresses this gap by integrating semantic hierarchies into the active learning pipeline for point cloud segmentation.

Hierarchical label structures have long been used in NLP and image understanding via manually curated ontologies like WordNet~\cite{miller1995wordnet} or domain standards such as CityGML~\cite{groger2012citygml}. More recent data-driven methods generate hierarchies through word embedding clustering or graph-based analysis~\cite{fu2014learning, sarfraz2022hierarchical}, though they often lack task adaptability. Large language models (LLMs) offer a new paradigm, with recent studies such as SHiNe~\cite{liu2024shine} leveraging GPT-based models to construct semantic hierarchies for open-vocabulary detection. Yet, the use of LLMs for building taxonomies in 3D segmentation remains largely unexplored. Concurrently, label hierarchies have been used to enhance segmentation performance via multi-level prediction~\cite{li2025deep}, hierarchical losses~\cite{camuffo2023learning}, and semantic consistency constraints~\cite{Mo_2019_CVPR}. Still, their potential for guiding active sample selection is untapped. Our work unifies these threads by introducing LLM-guided taxonomies and hierarchical uncertainty into the active learning loop, enabling more semantically meaningful and efficient annotation in 3D point cloud segmentation.

\begin{figure}
  \centering
  \includegraphics[width=\textwidth]{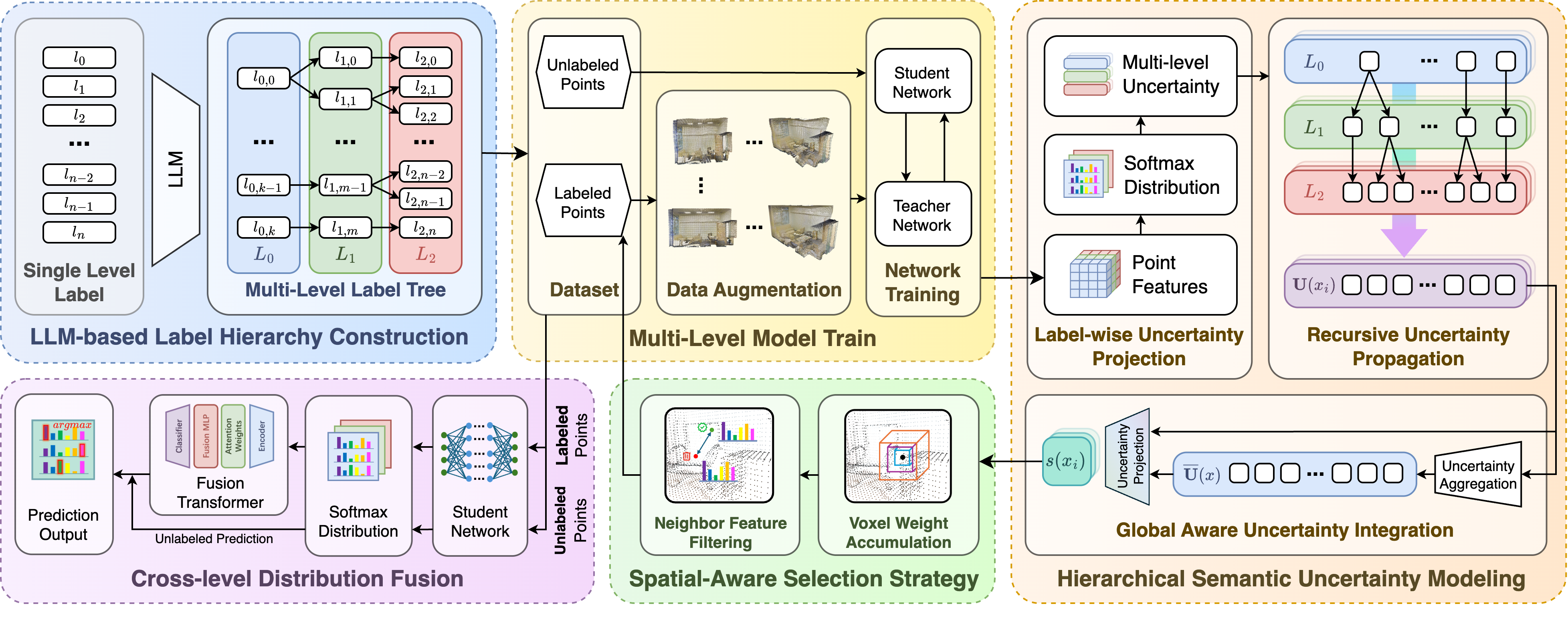}
  \caption{%
  \textbf{LLM-Guided Hierarchical-Label Uncertainty-Aware Active-Learning Framework. }%
    \textcolor[rgb]{0.424,0.557,0.749}{(a)~\textbf{LLM-based Hierarchical Label Construction}:} a single-level label set is partitioned by a large language model into a multi-level semantic tree, providing hierarchical priors for the subsequent modules; %
    \textcolor[rgb]{0.839,0.714,0.337}{(b)~\textbf{Multi-granular Prediction Training}:} softmax probability distributions at different semantic granularities are produced by a teacher–student network, laying the foundation for subsequent uncertainty estimation; %
    \textcolor[rgb]{0.600,0.396,0.000}{(c)~\textbf{Hierarchical Semantic Uncertainty Modelling}:} point features undergo \emph{Label-wise Projection}, followed by \emph{Recursive Propagation} and \emph{Global Integration} to obtain a cross-level uncertainty score; %
    \textcolor[rgb]{0.267,0.431,0.173}{(d)~\textbf{Spatial-aware Selection Strategy}:} voxel-wise aggregation and neighbourhood feature filtering are combined to choose diverse and representative query points according to the uncertainty ranking;%
    \textcolor[rgb]{0.584,0.467,0.639}{(e)~\textbf{Cross-level Distribution Fusion}:} during inference, softmax distributions from all semantic levels are fed into a lightweight attention module, which adaptively fuses them to yield the final prediction. %
    With only a few active-learning rounds, the framework markedly improves 3D point-cloud semantic segmentation under an extremely low annotation budget.%
  }
  \label{fig:framework}
  \vspace{-1em}
\end{figure}

\section{Methodology}

\subsection{Overview}
We propose a hierarchical label-aware semi-supervised active learning framework for efficient 3D point cloud segmentation (Figure~\ref{fig:framework}). Starting from single-level labels, we use a large language model (LLM) to construct a multi-level semantic hierarchy. With a small set of labeled points and abundant unlabeled data, we train a teacher-student network: the teacher generates pseudo-labels via EMA, and the student learns from both labeled and unlabeled data.  To guide point selection, we compute uncertainty scores across semantic levels using a top-down propagation strategy and enhance scoring with global uncertainty alignment. Final selection ranks are refined by incorporating voxel-level uncertainty and applying spatial redundancy filtering to ensure diversity. Newly labeled points are then used to update the model, and the process repeats until the annotation budget is met.  Each component of the framework is detailed in the following sections.

\subsection{LLM-based Label Hierarchy Construction}

To construct a hierarchical semantic label structure suitable for 3D point cloud segmentation, we initially explored traditional semantic embedding-based methods (e.g., Word2Vec clustering~\cite{mikolov2013efficient}). However, these methods rely solely on fixed semantic relationships defined by word embeddings, lacking customization for 3D scene recognition tasks, and thus cannot effectively capture the inherent semantic hierarchy present in real-world object categories. Additionally, the rigid nature of embedding-based clustering methods often results in unreasonable hierarchical groupings and ambiguity in label distinctions, particularly when handling semantically unclear labels such as ``other furniture,'' thereby limiting the potential for model accuracy improvements.

To address these limitations, we propose using Large Language Models (LLMs) to directly generate hierarchical label structures, thereby better aligning with task-specific requirements and human semantic understanding. Specifically, we employed a detailed prompt-engineering approach to guide LLMs in generating multi-level label trees, iteratively refining the prompts to progressively enhance the generated structures. Our prompt design incorporated the following key constraints:

\begin{itemize}
    \item \textbf{Distinct Hierarchical Separation}: Ensuring clear semantic boundaries between hierarchical levels to avoid semantic overlap and ambiguity.
    \item \textbf{Class Balance Control}: Maintaining a relatively balanced number of labels within each semantic level to avoid overly dense or sparse categories.
    \item \textbf{Flexible Handling of Ambiguous Labels}: Allowing semantically ambiguous labels (e.g., labels containing ``other'') to be categorized independently or flexibly, thereby enhancing the overall clarity and usability of the label hierarchy.
\end{itemize}

Through six iterations of prompt refinement, we systematically addressed specific issues identified in initial LLM-generated structures, such as duplicated labels, unclear hierarchical relationships, and difficulties in categorizing ambiguous labels. Eventually, rigorous human evaluations and comparative experiments demonstrated that the LLM-generated hierarchical label structures significantly outperformed traditional methods. In an anonymous A/B test involving 136 participants, approximately 77.2\% found that the LLM-generated hierarchy was superior in terms of semantic clarity, hierarchical coherence, and interpretability. Particularly, improvements were notable in handling ambiguous labels and achieving category balance compared to structures produced by the Word2Vec method. Consequently, we fully adopt the LLM-generated hierarchical label structure throughout this work.

\subsection{Hierarchical Semantic Uncertainty Modeling}

In 3D point cloud segmentation, fine-grained categories (e.g., \textit{chair}, \textit{toilet}) often have visually similar shapes and structures, making them difficult to distinguish from raw geometry alone. However, these classes typically belong to more semantically distinct coarse-level categories (e.g., \textit{furniture} vs. \textit{sanitary}). This motivates us to incorporate hierarchical label information and propagate uncertainty from coarse to fine levels, such that higher-level semantic differences can guide fine-grained decision-making.

\paragraph{Hierarchical Label Structure.} Let the label set $\mathcal{L}$ be organized into $n+1$ levels of increasing granularity, indexed from coarse to fine: $L_0, L_1, ..., L_n$. A label $l_{i,j} \in L_i$ represents the $j$-th label in level $i$. We define a parent-child relationship $l_{i,j} \preceq l_{i+1,k}$ if $l_{i,j}$ is a parent label of $l_{i+1,k}$ (or $l_{i+1,k}$ is a sublabel of $l_{i,j}$), where level $0$ is the coarsest and $n$ is the finest.

Each fine-grained label $l_{n,j} \in L_n$ defines a unique hierarchical label path:

\begin{equation}
\mathcal{P}_j = \{l_{0,j_0}, l_{1,j_1}, ..., l_{n,j_n}\}, \quad \text{with } l_{i,j_i} \preceq l_{i+1,j_{i+1}}.
\end{equation}

We denote the label $l$ at level $i$ in this path as $\mathcal{P}_{j,i}$.

For any label $\mathcal{P}_{j,i}$ with $i < n$, its sub-labels from the next finer level are:

\begin{equation}
\textbf{Sub}(\mathcal{P}_{j,i}) = \{ \mathcal{P}_{j,i+1} \in L_{i+1} \mid \mathcal{P}_{j,i} \preceq \mathcal{P}_{j,i+1} \}.
\end{equation}

\paragraph{Label-wise Uncertainty Projection.}
Given a point $x$, the predicted softmax distribution at level $i$ is:

\begin{equation}
P_i(x) = \{p_{l}(x)\ | l \in L_i \}
\end{equation}

where $p_l$ is the predicted probability of label $l$ for $x$. It is obvious that:

\begin{equation}
\sum_{l \in L_i} p_l(x) = 1
\end{equation}

We define the uncertainty of label $l$ at point $x$ as:

\begin{equation}
u_{l}(x) = 1 - 2|p_{l}(x) - 0.5|.
\end{equation}

\paragraph{Recursive Uncertainty Propagation.}

We define the propagated uncertainty $U_{\mathcal{P}_{j,i}}(x)$ as the cumulative uncertainty of label $\mathcal{P}_{j,i}$, integrating both its own uncertainty and the aggregated uncertainty from all coarser levels. The propagation starts from the coarsest level and proceeds layer by layer toward finer granularity.

The initial uncertainty is set as:

\begin{equation}
U_{\mathcal{P}_{j,0}}(x) = u_{\mathcal{P}_{j,0}}(x),
\end{equation}

Subsequent levels are computed recursively using:


\begin{equation}
\begin{aligned}
U_{\mathcal{P}_{j,i}}(x)
&= u_{\mathcal{P}_{j,i}}(x)
   + \omega \cdot U_{\mathcal{P}_{j,i-1}}(x) \cdot{}
\frac{u_{\mathcal{P}_{j,i}}(x)}
        {\sum_{\mathcal{P}_{k,i}\in \mathbf{Sub}(\mathcal{P}_{j,i-1})} u_{\mathcal{P}_{k,i}}(x)}
\end{aligned}
\label{eq:recursive_uncertainty}
\end{equation}

where $\omega$ is a layer-specific decay factor controlling the influence of coarser-level uncertainty on finer labels.

By integrating uncertainty across coarse-to-fine semantic levels, this propagation approach mitigates local ambiguity and enhances the discriminability of similar fine-grained categories.

\paragraph{Global Aware Uncertainty Integration.}

After recursively computing the propagated uncertainty across the hierarchy, we focus on the finest level, which reflects the cumulative uncertainty of each leaf label path. For a given point $x$, we collect the propagated uncertainties of all paths at the final level:

\begin{equation}
\mathbf{U}(x) = \left[ U_{\mathcal{P}_{0,n}}(x),\ U_{\mathcal{P}_{1,n}}(x),\ \dots,\ U_{\mathcal{P}_{m,n}}(x) \right],
\end{equation}

where $m = |L_n|$ is the number of fine-grained labels.

To model the global uncertainty distribution over the unlabeled points in the  cloud, we compute the mean uncertainty vector:

\begin{equation}
\overline{\mathbf{U}} = \frac{1}{N} \sum_{i=1}^{N} \mathbf{U}(x_i),
\end{equation}

where $x_1, ..., x_N$ are the unlabeled points. This mean vector reflects the model's uncertainty tendencies across the fine-grained label space.

Based on this, we define the uncertainty score of each point as the inner product between its uncertainty vector and the global profile:

\begin{equation}
s(x_i) = \overline{\mathbf{U}} \cdot \mathbf{U}(x_i)^\top.
\end{equation}

This formulation favors points that exhibit high uncertainty while aligning with the global uncertainty pattern, thereby prioritizing samples that are not only informative and structurally representative, but also more likely to improve model generalization when annotated.





\subsection{Spatial-Aware Selection Strategy} 
We adopt the HPAL~\cite{xu2023hierarchical} framework, incorporating voxel-based sampling, a teacher-student (TS) semi-supervised training scheme, and the Feature Distance Suppression (FDS) module. The point cloud is divided into voxels, from which the highest-uncertainty ones are selected by aggregating point-level scores; a fixed number of points are then sampled per voxel to ensure spatial coverage. FDS further filters out feature-redundant neighbors to enhance diversity. The student model is trained using both labeled and pseudo-labeled data, while the teacher is updated via EMA. Although the core training structure aligns with HPAL, our method is guided by cross-hierarchy uncertainty, introducing stronger semantic priors into the point selection process.

\subsection{Cross-level Distribution Fusion}
To leverage semantic information across multiple granularity levels, we introduce a cross-level distribution fusion module. For each point, probability distributions from different semantic levels are projected into a shared hidden space using level-specific linear encoders with ReLU activations:
\begin{equation}
h_i = \phi_i(p_i) = \mathrm{ReLU}(W_i p_i + b_i)
\end{equation}
An attention mechanism then learns the importance of each level by computing normalized weights:
\begin{equation}
\alpha_i = \frac{\exp(W_a h_i)}{\sum_{j} \exp(W_a h_j)}
\end{equation}
The fused representation is obtained via weighted summation of encoded features, followed by a fusion MLP and a final classifier that outputs fine-grained predictions. This approach dynamically integrates multi-level semantic cues and outperforms simple fusion strategies like averaging or fixed-weight voting.

\section{Experiments}

\subsection{Datasets}

We evaluate our method on two widely-used 3D indoor segmentation benchmarks: S3DIS and ScanNet v2, covering diverse scene types and label distributions.

\textbf{S3DIS} contains 6 building areas with 271 rooms. We follow the standard protocol using Area 5 for testing and construct a 3-level semantic hierarchy with 3, 6, and 13 classes.

\textbf{ScanNet v2} includes over 1500 RGB-D scenes. We adopt the official split and define a 3-level label structure with 3, 9, and 20 classes.

Hierarchical labels are generated using both Word2Vec-based clustering and LLM-guided prompting (e.g., GPT-4o/GPT-o3). An A/B test with 136 participants showed strong preference (77.2\%) for the LLM-derived hierarchies, which are adopted in our experiments. Details are provided in the Appendix.

\begin{figure}[!t]
    \centering
    \captionsetup[subfigure]{skip=2pt}
    \begin{subfigure}[b]{0.195\textwidth}
        \includegraphics[width=\textwidth]{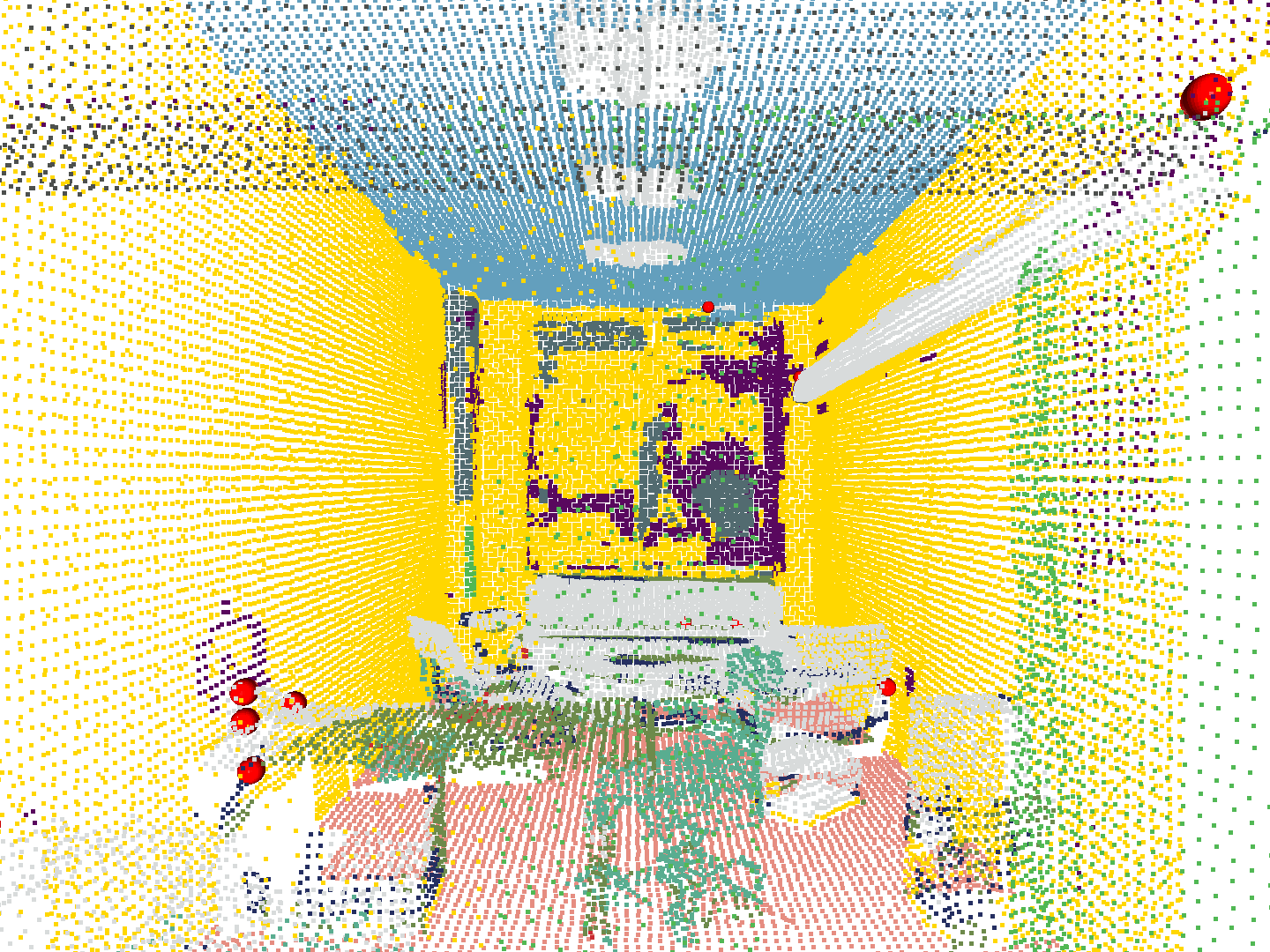}
        \caption*{Iteration 0}
    \end{subfigure}
    \begin{subfigure}[b]{0.195\textwidth}
        \includegraphics[width=\textwidth]{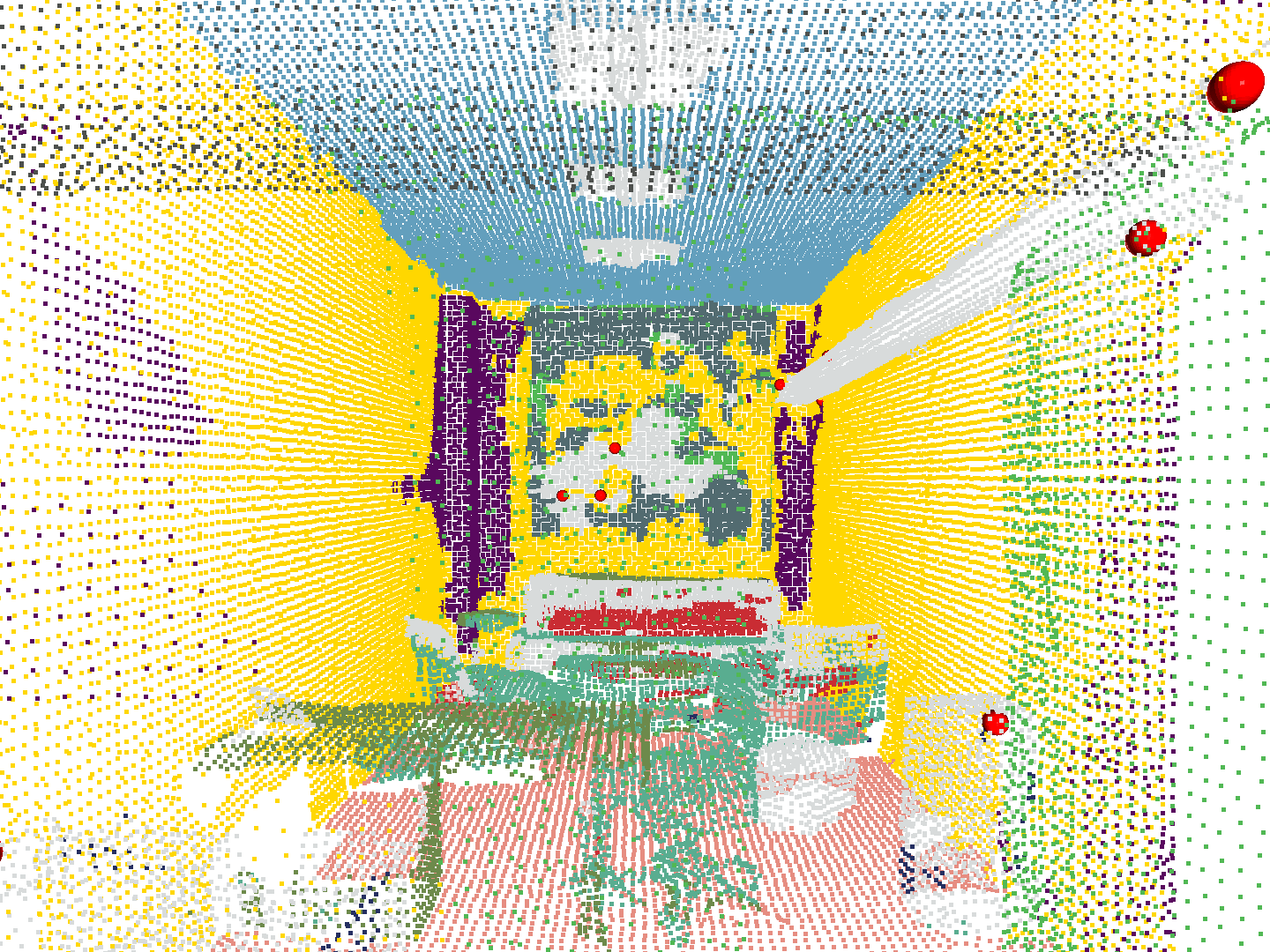}
        \caption*{Iteration 1}
    \end{subfigure}
    \begin{subfigure}[b]{0.195\textwidth}
        \includegraphics[width=\textwidth]{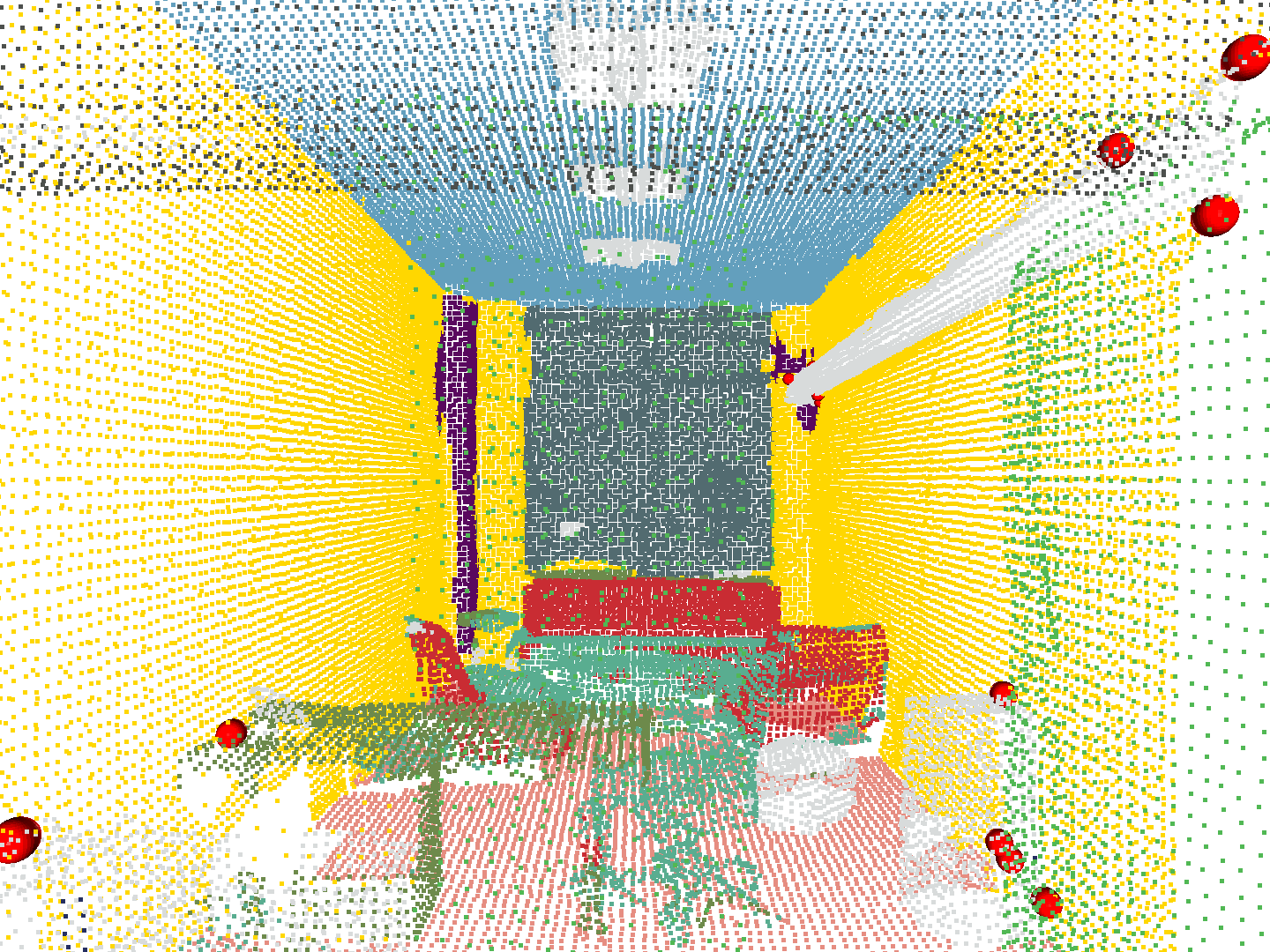}
        \caption*{Iteration 2}
    \end{subfigure}
    \begin{subfigure}[b]{0.195\textwidth}
        \includegraphics[width=\textwidth]{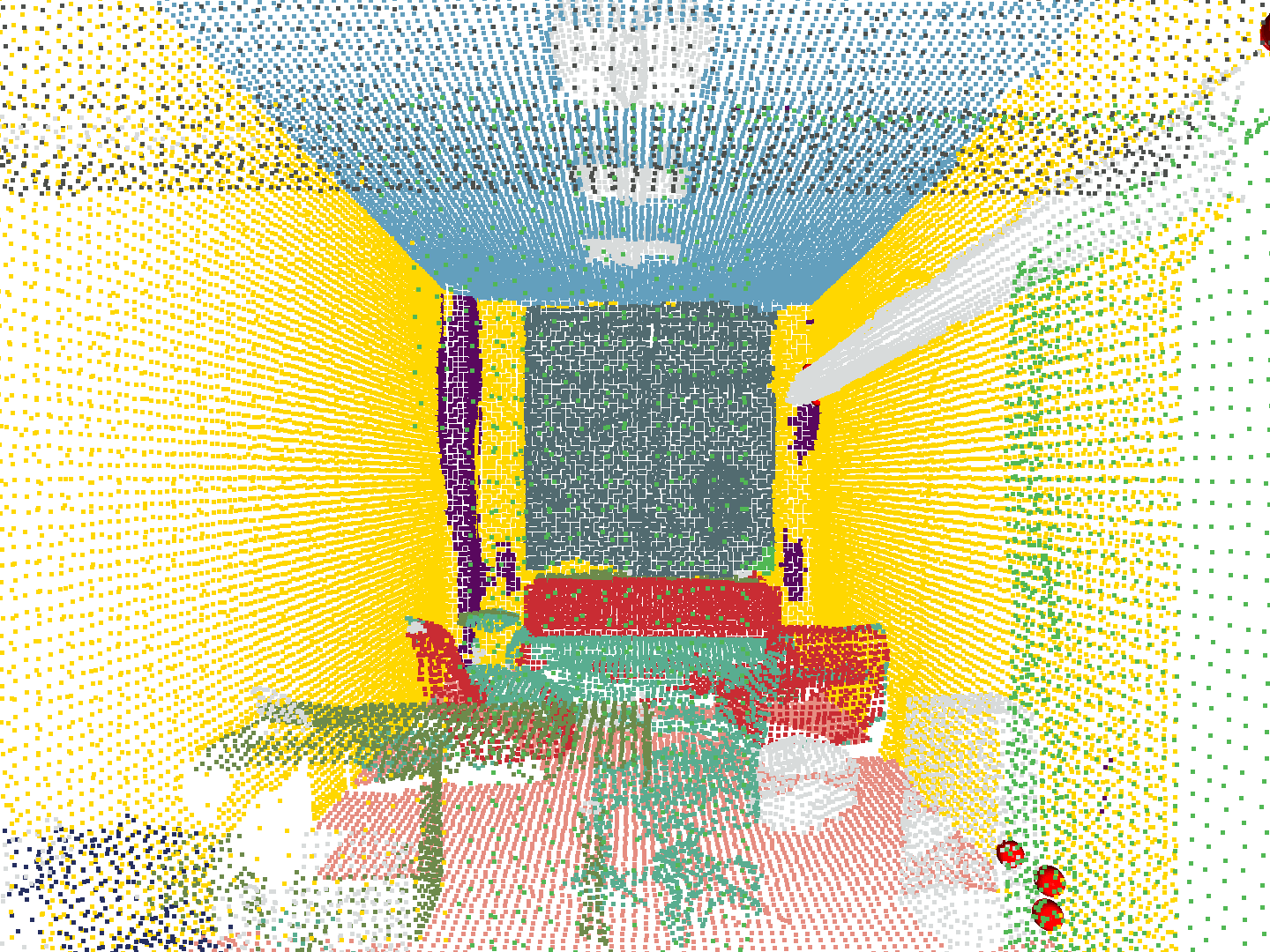}
        \caption*{Iteration 3}
    \end{subfigure}
    \begin{subfigure}[b]{0.195\textwidth}
        \includegraphics[width=\textwidth]{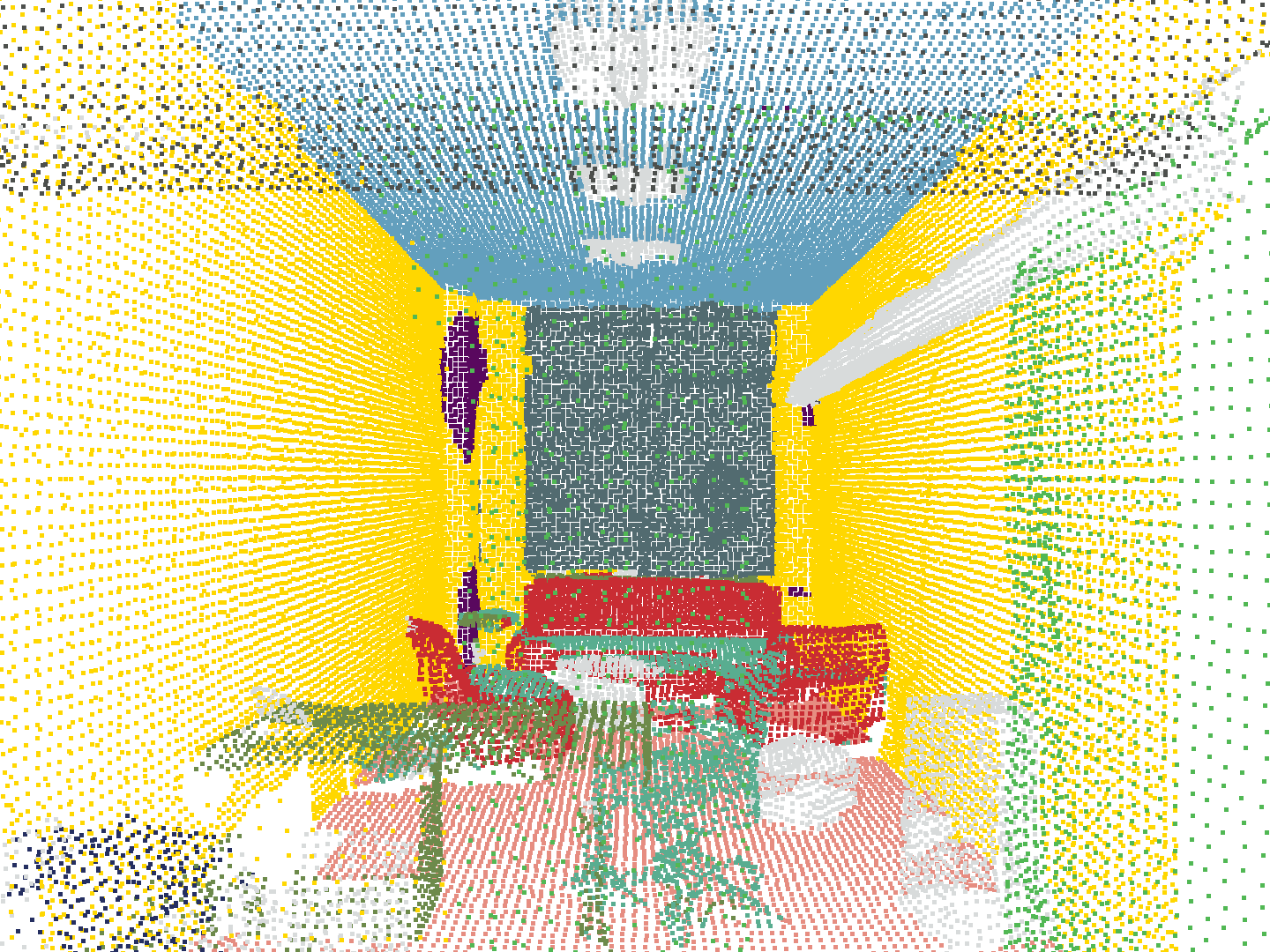}
        \caption*{iteration 4}
    \end{subfigure}%
    
    \definecolor{ceiling}{rgb}{0.388, 0.624, 0.741}
    \definecolor{floor}{rgb}{0.902, 0.553, 0.506}
    \definecolor{wall}{rgb}{1.000, 0.843, 0.000}
    \definecolor{beam}{rgb}{0.302, 0.314, 0.302}
    \definecolor{column}{rgb}{0.349, 0.031, 0.369}
    \definecolor{window}{rgb}{0.322, 0.420, 0.439}
    \definecolor{door}{rgb}{0.318, 0.722, 0.329}
    \definecolor{chair}{rgb}{0.349, 0.678, 0.565}
    \definecolor{table}{rgb}{0.427, 0.541, 0.294}
    \definecolor{bookcase}{rgb}{0.145, 0.184, 0.380}
    \definecolor{sofa}{rgb}{0.788, 0.173, 0.200}
    \definecolor{board}{rgb}{0.800, 0.749, 0.933}
    \definecolor{clutter}{rgb}{0.847, 0.859, 0.859}
    \newcommand{\legenditem}[2]{%
      \raisebox{0pt}[0pt][0pt]{%
        \begin{tikzpicture}[baseline=0.35ex]
          \filldraw[draw=none,fill=#1] (0,0) rectangle +(0.3,0.3);
        \end{tikzpicture}}%
      \hspace{0.2em}#2
    }
    \vspace{-0.5em}
    \begin{center}
    {\footnotesize
      \legenditem{ceiling}{ceiling}\hspace{0.1em}
      \legenditem{floor}{floor}\hspace{0.1em}
      \legenditem{wall}{wall}\hspace{0.1em}
      \legenditem{beam}{beam}\hspace{0.1em}
      \legenditem{column}{column}\hspace{0.1em}
      \legenditem{window}{window}\hspace{0.1em}
      \legenditem{door}{door}\\[0.1em]
      \legenditem{chair}{chair}\hspace{0.1em}
      \legenditem{table}{table}\hspace{0.1em}
      \legenditem{bookcase}{bookcase}\hspace{0.1em}
      \legenditem{sofa}{sofa}\hspace{0.1em}
      \legenditem{board}{board}\hspace{0.1em}
      \legenditem{clutter}{clutter}
    }
    \end{center}

    \vspace{-0.5em}
    \captionsetup{font=footnotesize}
    \caption{Visualization of five rounds of semantic predictions and point‑selection results for the office\_2 scene in Area 1 of the S3DIS dataset. The points selected in each iteration significantly refine the predictions of neighboring regions in the next iteration.}

  \label{fig:office2_iterations}
  \vspace{0.5em}
  \centering
  \captionsetup[subfigure]{font=footnotesize, skip=2pt}
  \captionsetup{font=footnotesize}

  \begin{subfigure}[b]{0.21\textwidth}
    \includegraphics[width=\textwidth]{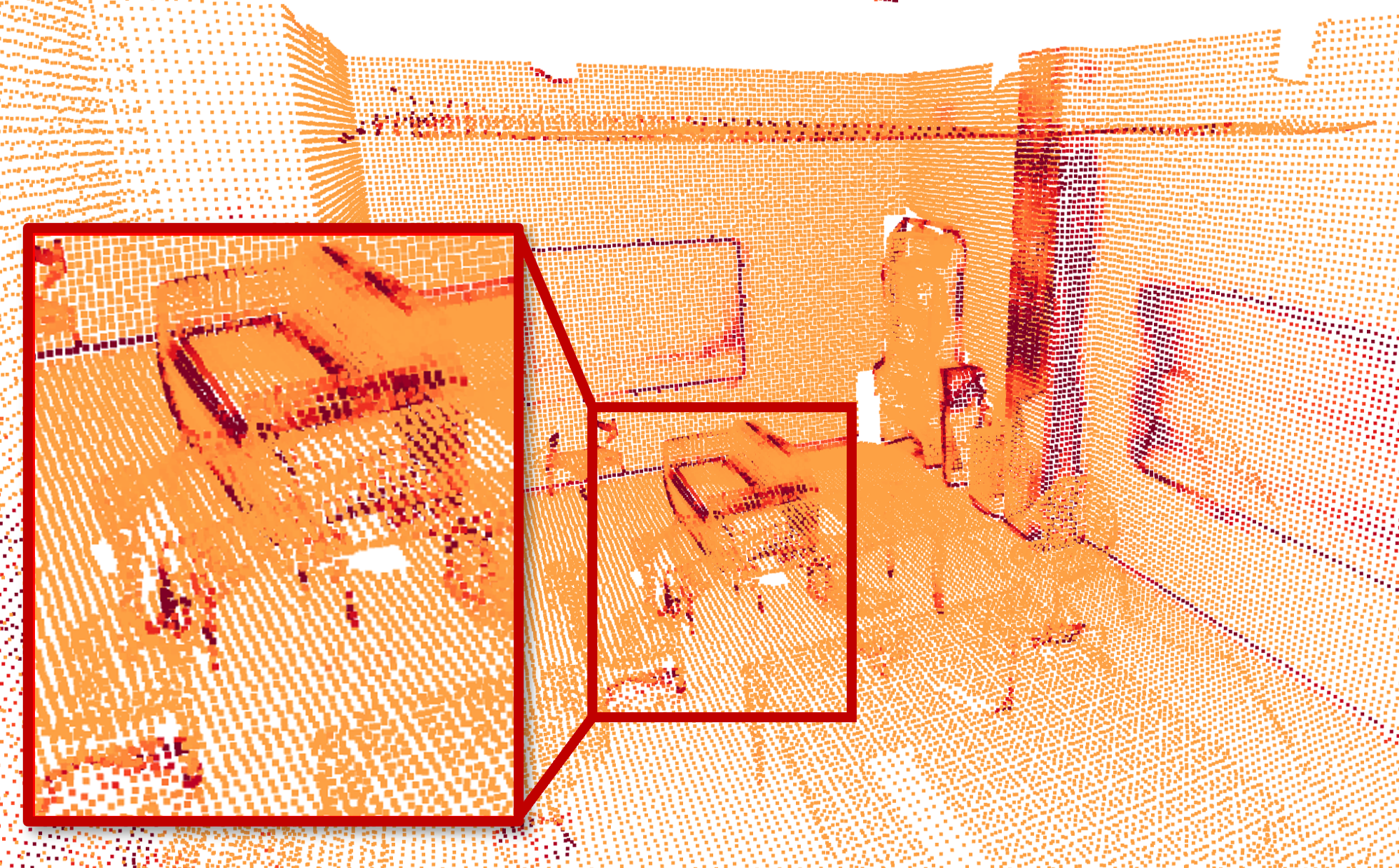}
    \caption*{(a)Baseline }
  \end{subfigure}
  \begin{subfigure}[b]{0.21\textwidth}
    \includegraphics[width=\textwidth]{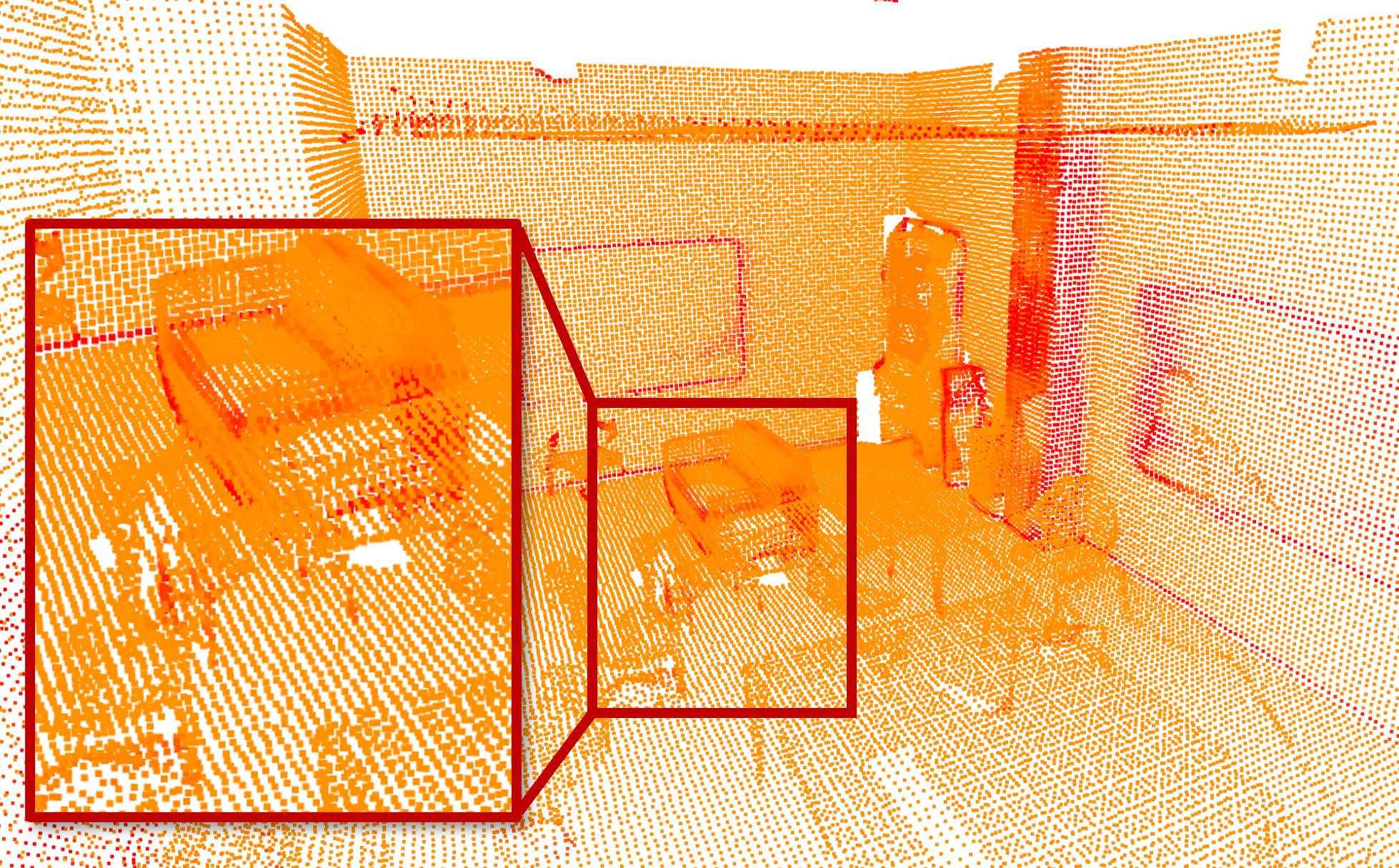}
    \caption*{(b) Our method}
  \end{subfigure}
  \begin{subfigure}[b]{0.21\textwidth}
    \includegraphics[width=\textwidth]{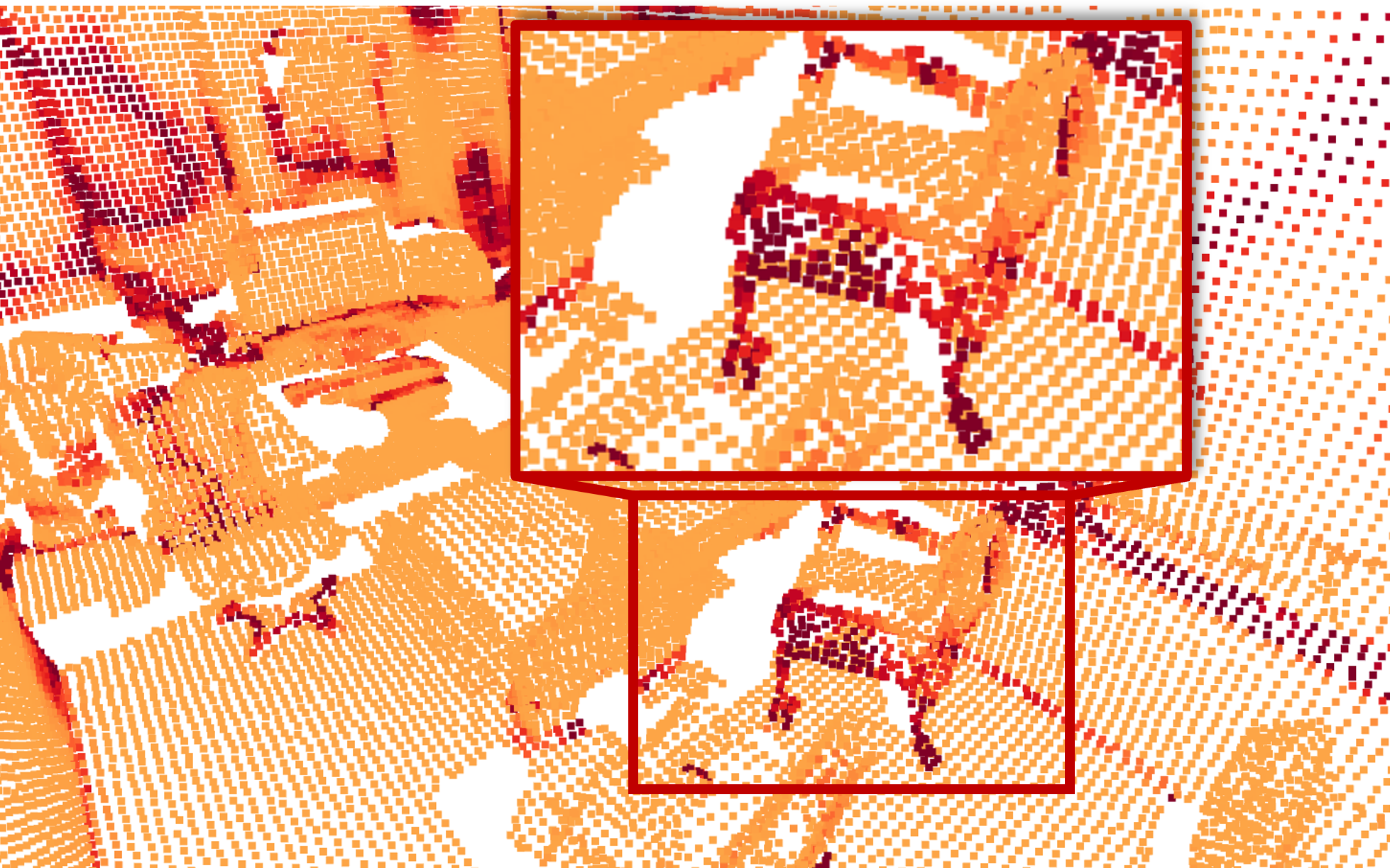}
    \caption*{(c) Baseline}
  \end{subfigure}
  \begin{subfigure}[b]{0.21\textwidth}
    \includegraphics[width=\textwidth]{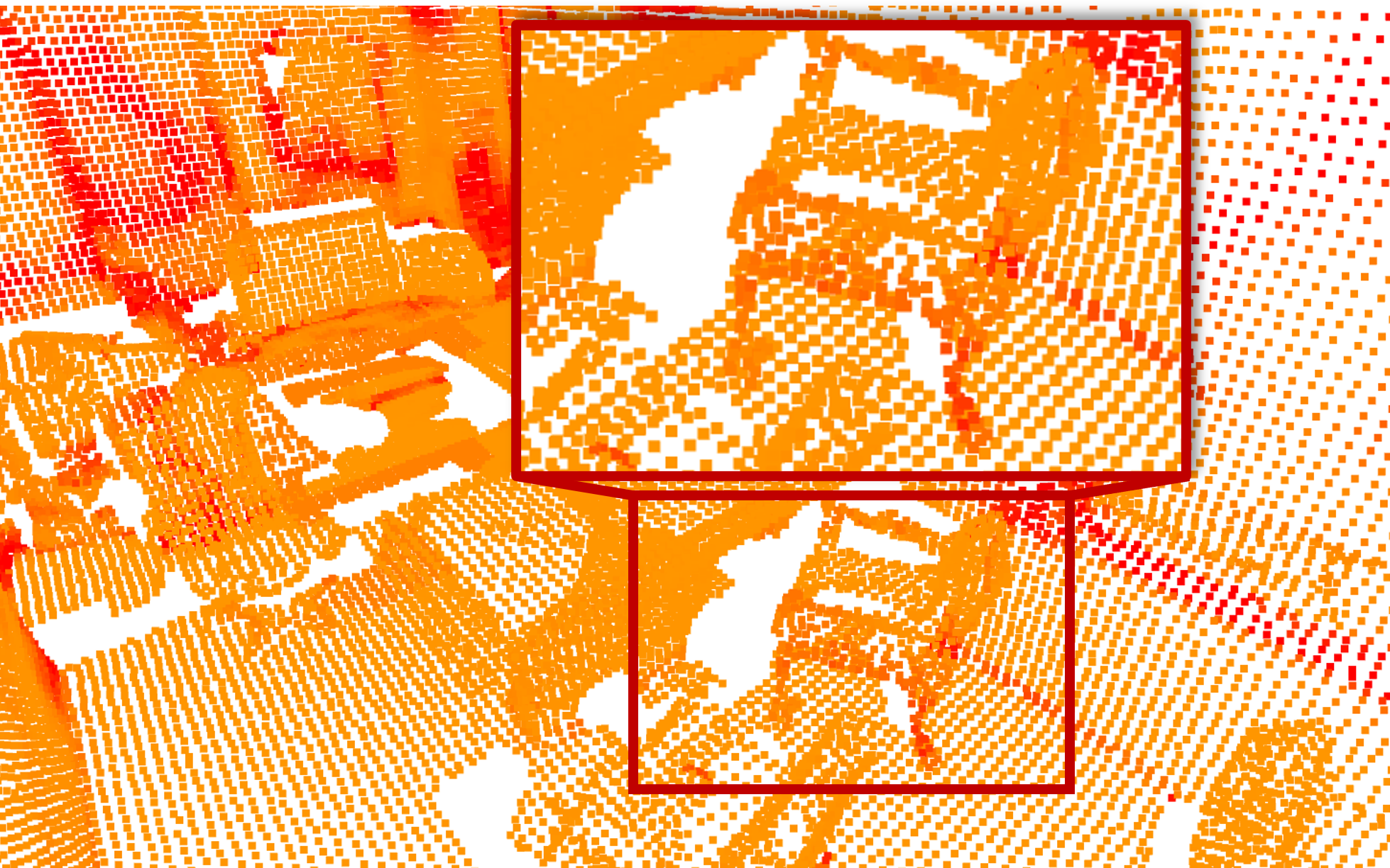}
    \caption*{(d) Our method}
  \end{subfigure}
  \vspace{-0.5em}
  \caption{Uncertainty heat‑map comparisons of two scenes of the S3DIS training set. Darker colours denote higher prediction uncertainty. Overall, our method produces smoother uncertainty estimates and yields better performance in complex regions such as object boundaries.}
  \label{fig:uncert_heatmaps_flat}
\end{figure}

\subsection{Implementation Details}

We implement our method using MinkUNet from MinkowskiEngine, employing sparse voxel convolutions to extract 3D features. All models are trained with SGD (initial learning rate 0.1) under a teacher-student semi-supervised framework, where the teacher is updated via EMA (rate 0.955) and pseudo-labels are filtered with a confidence threshold of 0.75.  Both datasets are trained for 60,000 steps using a batch size of 2 and cosine annealing for learning rate scheduling. S3DIS point clouds are voxelized to 0.04m, while ScanNet v2 uses original resolution.  We perform 5 active learning rounds, labeling the top 0.02\% most uncertain points per round using a hierarchical uncertainty fusion mechanism (decay factor $\omega=0.1$). To enhance spatial and semantic diversity, we adopt multi-scale voxelization and the FDS module~\cite{xu2023hierarchical}. The process is parallelized across 3 GPUs and optimized for efficient scoring. Each selection iteration completes in under 5 minutes. Additional details are deferred to the Appendix.  

\subsection{Result Analysis} 
To validate the effectiveness of our proposed framework, we compare our full method (denoted as Ours) against the baseline model, which shares the same backbone but is trained without any of our proposed modules, including the prediction refinement and label hierarchy modeling. All experiments are conducted on the S3DIS dataset, and the results on Area 5 are summarized in Table \ref{tab:weakly-supervised methods on S3DIS dataset}.

\begin{table*}[h]
\centering
\begin{minipage}[t]{0.48\textwidth}
    \scriptsize
    \centering
    \renewcommand{\arraystretch}{0.95}
    \setlength{\tabcolsep}{4pt}
    \resizebox{\linewidth}{!}{%
    \begin{tabular}{c|c|c}
        \hline
        Methods & Labeled points & Area 5 \\
        \hline\hline
        PointNet \cite{qi2017pointnet} & 100\% & 41.1 \\
        KPConv \cite{thomas2019kpconv} & 100\% & 67.1 \\
        MinkowskiNet \cite{choy20194d} & 100\% & 64.5 \\
        PointTransformerV3 \cite{Wu_2024_CVPR} & 100\% & 74.7 \\
        Swin3D \cite{yang2025swin3d} & 100\% & 72.5 \\
        \hline
        PSD \cite{zhang2021perturbed} & 1\% & 63.5 \\
        SQN \cite{hu2022sqn} & 1\% & 63.7 \\
        HybridCR \cite{li2022hybridcr} & 1\% & 65.3 \\
        \hline
        WSSS-ST \cite{yin2023label} & 0.1\% & 59.2 \\
        SQN \cite{hu2022sqn} & 0.1\% & 61.4 \\
        DR-Net \cite{zhang2024weakly} & 0.1\% & 58.7 \\
        Baseline & 0.1\% & 55.08 \\
        Ours & 0.1\% & \textbf{62.25} \\
        \hline
        PSD \cite{zhang2021perturbed} & 0.03\% & 48.2 \\
        HybridCR \cite{li2022hybridcr} & 0.03\% & 51.5 \\
        MIL \cite{yang2022mil} & 0.02\% & 51.4 \\
        GaIA \cite{lee2023gaia} & 0.02\% & 53.7 \\
        Baseline & 0.02\% & 49.81 \\
        Ours & 0.02\% & \textbf{54.03} \\
        \hline
    \end{tabular}
    }
    \captionsetup{justification=centering}
    \caption{Comparison with weakly-supervised methods on S3DIS Area-5}
    \label{tab:weakly-supervised methods on S3DIS dataset}
\end{minipage}
\hfill
\begin{minipage}[t]{0.48\textwidth}
    \scriptsize
    \centering
    \renewcommand{\arraystretch}{0.95}
    \setlength{\tabcolsep}{3pt}
    \resizebox{\linewidth}{!}{%
    \begin{tabular}{c|c|cc}
        \hline
        Methods & Labeled points & val & test \\
        \hline\hline
        PointNet++ \cite{qi2017pointnet++} & 100\% & – & 33.9 \\
        MinkowskiNet \cite{choy20194d} & 100\% & – & 72.1 \\
        BPNet \cite{hu2021bidirectional} & 100\% & – & 74.9 \\
        PTv3 \cite{Wu_2024_CVPR} & 100\% & 78.6 & 79.4 \\
        OA-CNNs \cite{Peng_2024_CVPR} & 100\% & 76.1 & 75.6 \\
        \hline
        SPG \cite{deng2022superpoint} & 10\% & – & 52.38 \\
        WS3D \cite{liu2022weakly} & 10\% & 59.3 & – \\
        \hline
        PSD \cite{zhang2021perturbed} & 1\% & – & 54.7 \\
        HybridCR \cite{li2022hybridcr} & 1\% & 56.9 & 56.8 \\
        GaIA \cite{lee2023gaia} & 1\% & 65.2 \\
        \hline
        SQN \cite{hu2022sqn} & 0.1\% & – & 56.9 \\
        Ours & 0.1\% & 63.45 & – \\
        \hline
        SQN \cite{hu2022sqn} & 20pts & – & 48.6 \\
        PointContrast \cite{xie2020pointcontrast} & 20pts & – & 55.0 \\
        OTOC \cite{Liu_2021_CVPR} & 20pts & 61.35 & 59.4 \\
        MIL \cite{yang2022mil} & 20pts & 57.8 & 54.4 \\
        VIBUS \cite{tian2022vibus} & 20pts & – & 58.6 \\
        Ours & 20pts & 61.05 & – \\
        \hline
    \end{tabular}
    }
    \captionsetup{justification=centering}
    \caption{Comparison with weakly-supervised methods on ScanNet v2}
    \label{tab:weakly-supervised methods on ScanNet dataset}
\end{minipage}
\end{table*}

As shown in Table \ref{tab:weakly-supervised methods on S3DIS dataset} for the S3DIS Area-5 dataset, our method achieves a mean mIoU of 62.25\% under the 0.1\% annotation budget, outperforming the baseline by 7.17\%. Notably, even with the minimal 0.02\% labeling budget, our framework still achieves 54.03\% (exceeding the baseline by 4.22\%), demonstrating its robustness under extremely low supervision. This significant improvement stems from our novel unified training framework, detailed in Figure \ref{fig:framework}, unlike the baseline lacking structural modeling or uncertainty guidance, integrates key advancements. Specifically, our full model leverages a Label Hierarchy Tree constructed by an LLM, a multi-level uncertainty modeling module, and a cross-level prediction fusion mechanism. Together, these components are crucial for integrating hierarchical semantics, sophisticated uncertainty modeling, and spatially-aware selection, enabling the model to better capture coarse-to-fine semantic cues and actively select the most informative unlabeled points for annotation.

As shown in the Table \ref{tab:weakly-supervised methods on ScanNet dataset}, for the ScanNet v2 dataset, our method's performance was assessed using two sparse annotation budgets: 0.1\% of total labeled points and a fixed 20-point budget per scene. With merely 0.1\% labels, our approach yielded a robust validation mIoU of 63.45\%. This result is notably higher than SQN which achieved 56.9\% using the same label percentage, and it also surpasses methods like HybridCR which reported 56.9\% even with a significantly larger 1\% annotation budget. In the extremely challenging 20-point per scene scenario, our method obtained a validation mIoU of 61.05\%, proving highly competitive with OTOC and outperforming others such as MIL. These strong validation outcomes across demanding, low-label settings effectively demonstrate our hierarchical active learning strategy's ability to identify crucial data points, thereby substantially reducing annotation efforts while preserving high segmentation performance.

We argue that this training paradigm is generalizable to other segmentation backbones. Since our design is modular and does not modify the core architecture, it can be easily integrated with recent point cloud encoders like PointTransformer or KPConv. We expect that similar gains could be achieved across backbones, thanks to the additional semantic priors and uncertainty-driven selection introduced by our framework.

\subsection{Ablation Studies}

We conduct ablation studies on S3DIS to evaluate the contributions of four core modules: LLM-based Label Hierarchy Construction (LLHC), Label-wise Uncertainty Projection (LUP), Recursive Uncertainty Propagation (RUP), and Global-Aware Uncertainty Integration (GAUI).

\begin{figure}[h]
\centering
\begin{minipage}[t]{0.49\textwidth}
    \centering
    \captionsetup{justification=centering}
    \includegraphics[width=\linewidth]{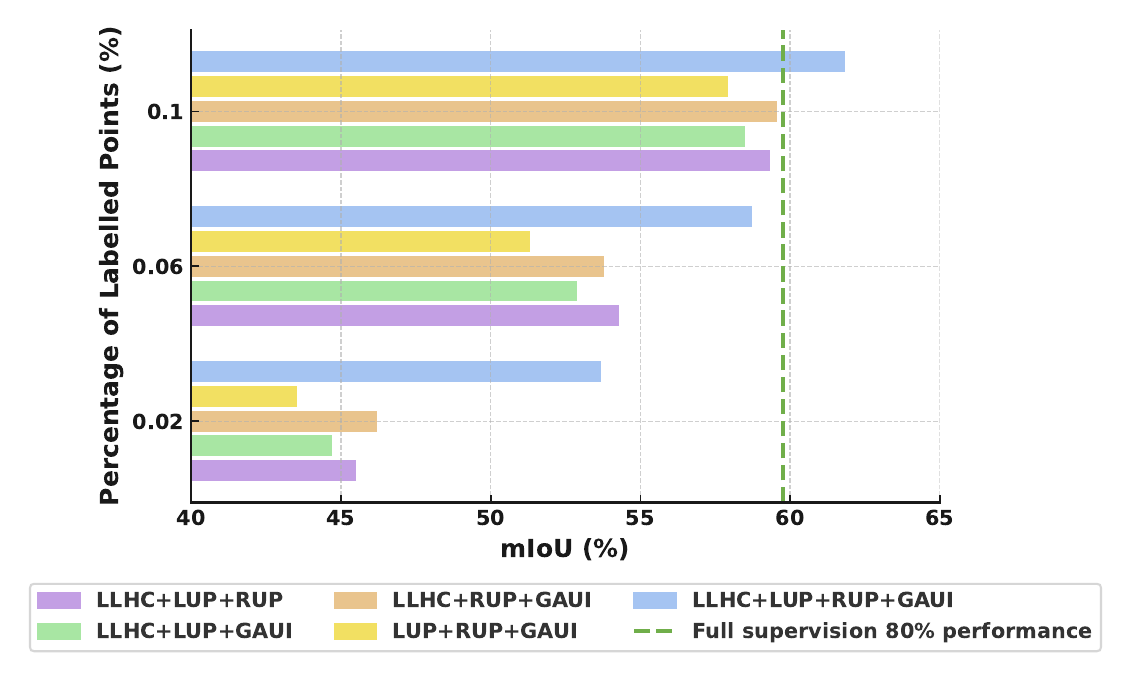}
    \caption{Ablation of module combinations under varying label ratios}
    \label{fig:ablation_study_custom}
\end{minipage}
\hfill
\begin{minipage}[t]{0.49\textwidth}
    \centering
    \captionsetup{justification=centering}
    \includegraphics[width=\linewidth]{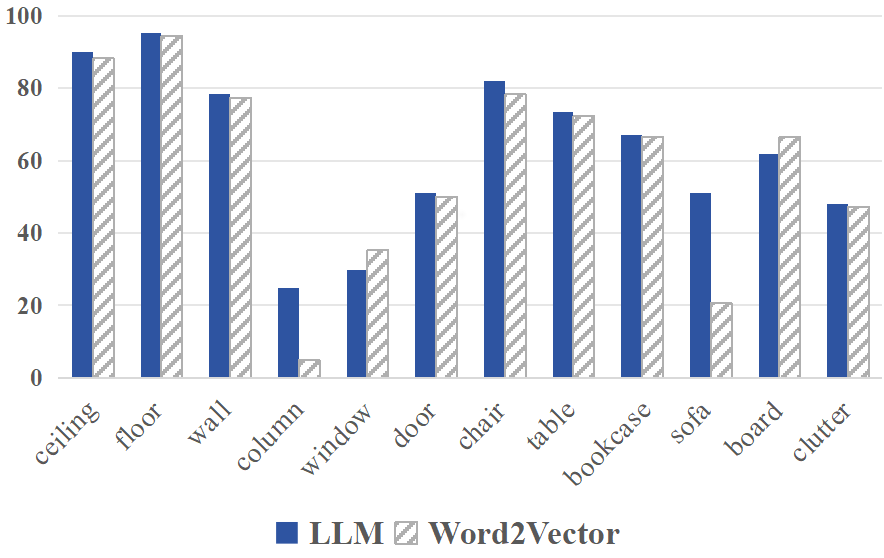}
    \caption{LLM versus Word2Vec for label hierarchy construction}
    \label{fig:llm_vs_w2v}
\end{minipage}
\vspace{-1em}
\end{figure}

To illustrate the impact of different modules, Figure~\ref{fig:ablation_study_custom} shows mIoU curves for five module combinations across annotation budgets of 0.02\%, 0.06\%, and 0.10\%. The full configuration (LLHC + LUP + RUP + GAUI) achieves the highest performance, reaching 62.25\% mIoU at 0.1\%, exceeding 80\% of the fully supervised baseline. Removing GAUI drops performance to 59.34\% (-2.91\%), and excluding RUP further lowers it to 58.48\%, highlighting the importance of recursive uncertainty propagation. The setup without LUP achieves 59.57\%, still below the full model. Notably, omitting LLHC leads to the worst performance (57.94\%), confirming the key role of LLM-guided hierarchical labels.

As shown in Table \ref{tab:ablation_hls_uem_pd_auw}, we specifically examined the contributions of four key modules: (1) \textbf{LLM-based Label Hierarchy Construction (LLHC)}, utilizing a three-level label hierarchy generated by a large language model to enhance semantic discrimination; (2) \textbf{Label-wise Uncertainty Projection (LUP)}, designed to effectively identify the most uncertain samples; (3) \textbf{Recursive Uncertainty Propagation (RUP)}, aiming to reduce redundant selections across iterations; and (4) \textbf{Global Aware Uncertainty Integration (GAUI)}, stabilizing the selection of points by integrating the model’s uncertainty scores. If the LLHC module is not used, the hierarchical labels generated by LLM are replaced with labels derived from Word2Vec clustering. Similarly, if the LUP module is not employed, uncertainty estimation will revert to using the traditional softmax distribution. The experimental results indicate that removing the hierarchical labels generated by the large language model caused the largest decrease of 4.31 mIoU. Removing the LUP module reduced performance by 2.68 mIoU, while removing the RUP module resulted in the performance drop of 3.77 mIoU. Omitting the GAUI module caused a decrease of 2.91 mIoU. These results demonstrate that all four modules substantially contribute to overall model performance, with the hierarchical label structure generated by the large language model and recursive uncertainty propagation having the most significant impacts.

\begin{table}[h]
    \scriptsize
    \centering
    \resizebox{0.5\linewidth}{!}{
    \begin{tabular}{c|c|c|c|c}
        \hline
        LLHC & LUP & RUP & GAUI & mIoU(\%) \\
        \hline\hline
         & $\checkmark$ & $\checkmark$ & $\checkmark$ & 57.94 \\
        $\checkmark$ &  & $\checkmark$ & $\checkmark$ & 59.57 \\
        $\checkmark$ & $\checkmark$ &  & $\checkmark$ & 58.48 \\
        $\checkmark$ & $\checkmark$ & $\checkmark$ &  & 59.34 \\
        $\checkmark$ & $\checkmark$ & $\checkmark$ & $\checkmark$ & 62.25 \\
        \hline
    \end{tabular}
    }
    \vspace{1ex}
    \caption{Ablation study of different module}
    \label{tab:ablation_hls_uem_pd_auw}
\end{table}

Furthermore, in Figure \ref{fig:llm_vs_w2v}, we compared the three-level hierarchical label structure generated by LLM with the traditional Word2Vec clustering approach. The experimental results indicate that the LLM-based approach outperformed the Word2Vec method in 10 out of 12 categories, significantly improving performance by approximately 20\% in the column category and 30\% in the sofa category. In general, most semantic labels achieved significant mIoU improvements using the hierarchical structure generated by LLM compared to the Word2Vec scheme, thus validating the effectiveness of LLM in semantic clustering tasks.

To further investigate the impact of different hierarchical configurations, we performed a series of ablation studies using semantic hierarchies of two layers, three layers, and four layers, the results summarized in Table~\ref{tab:layer_ablation}. The experimental results show that model performance steadily improves from 2-layer to 3-layer configurations, with the 3-layer hierarchy achieving the best overall result. Specifically, the 3-layer model reaches a final mIoU of 62.25\%, outperforming the 2-layer and 4-layer variants by 5.63\% and 2.74\%, respectively. Although the 4-layer configuration continues to show slight improvements in intermediate iterations, it does not surpass the 3-layer setting in the final result, possibly due to oversmoothing or increased difficulty in effective label propagation. Moreover, deeper hierarchies introduce additional overhead in memory and computation, particularly during uncertainty propagation and hierarchical fusion. Taking into account both performance and efficiency, we adopt the three-layer hierarchy as our final design choice.

\begin{table}[h]
\centering
\small
\begin{minipage}[t]{0.6\textwidth}
\centering
\resizebox{\textwidth}{!}{
\begin{tabular}{l|ccccc|c}
\hline
\textbf{Layers} & \textbf{iter1} & \textbf{iter2} & \textbf{iter3} & \textbf{iter4} & \textbf{iter5} & \textbf{Final mIoU (\%)} \\
\hline
2-layer & 36.16 & 44.29  & 51.70 & 53.34 & 54.61 & 56.62 \\
3-layer & 38.65 & 46.81 & 53.77 & 57.47 & 60.05 & \textbf{62.25} \\
4-layer & 37.78 & 45.86 & 52.91 & 56.27 & 57.84 & 59.51 \\
\hline
\end{tabular}
}
\captionsetup{justification=centering}
\vspace{1ex}
\caption{Ablation study of different hierarchical layer configurations.}
\label{tab:layer_ablation}
\end{minipage}
\hfill
\begin{minipage}[t]{0.35\textwidth}
\centering
\resizebox{\textwidth}{!}{
\begin{tabular}{l|c}
\hline
\textbf{Fusion Method} & \textbf{mIoU (\%)} \\
\hline
Weighted Addition & 59.56 \\
Simple Addition & 60.05 \\
Enhanced Attention Fusion & \textbf{62.25} \\
\hline
\end{tabular}
}
\captionsetup{justification=centering}
\vspace{1ex}
\caption{Comparison of different semantic fusion strategies.}
\label{tab:fusion_ablation}
\end{minipage}
\vspace{-1.0em}
\end{table}

To assess the impact of different semantic-level fusion strategies on the final prediction, we compare three approaches: direct addition, weighted addition, and an enhanced attention-based fusion module. As shown in Table~\ref{tab:fusion_ablation}, the attention-based fusion method achieves the highest performance, reaching 62.25\% mIoU. This validates its ability to adaptively learn the importance of each semantic level and assign dynamic weights accordingly. In contrast, the simple addition strategy yields 60.05\% mIoU, indicating that although semantic aggregation helps, equal weighting does not fully exploit the multi-level semantic information. The weighted addition strategy, which introduces predefined fixed weights for each level, results in 59.56\% mIoU. Despite being more flexible than equal addition, it still lacks the adaptability and contextual awareness of the learned attention mechanism. 

\section{Conclusion}
We propose a hierarchical label-aware active learning framework for 3D point cloud segmentation, addressing the limitations of flat label spaces. Our method combines LLM-guided taxonomy construction with a hierarchical uncertainty model to guide informative and semantically consistent point selection. Integrated with semi-supervised learning and spatial awareness, it achieves state-of-the-art performance on S3DIS and ScanNet v2 under low annotation budgets. Future work will explore deeper LLM-taxonomy integration and refined uncertainty modeling.


\bibliographystyle{unsrt}
\bibliography{LLM_Guided}

\end{document}